\documentclass[10pt,twocolumn,letterpaper]{article}

\usepackage[pagenumbers]{cvpr} % To force page numbers, e.g. for an arXiv version

\usepackage{graphicx}
\usepackage{amsmath}
\usepackage{amssymb}
\usepackage{booktabs}
\usepackage{times}
\usepackage{epsfig}
\usepackage{upgreek}
\usepackage{appendix}
\usepackage{diagbox}
\usepackage{multirow}
\usepackage{times}
\usepackage{diagbox}
\usepackage{bm}
\usepackage{color}
\usepackage{makecell}

\newcommand{\tabincell}[2]{\begin{tabular}{@{}#1@{}}#2\end{tabular}}

\usepackage[pagebackref,breaklinks,colorlinks]{hyperref}
\usepackage[accsupp]{axessibility} % Improves PDF readability for those with visual impairments.

\graphicspath{{figs/}}

\makeatletter
\def\thanks#1{\protected@xdef\@thanks{\@thanks
        \protect\footnotetext{#1}}}
\makeatother

\begin{document}

\title{Unsigned Orthogonal Distance Fields: An Accurate Neural Implicit Representation for Diverse 3D Shapes\vspace{-4mm}}

\author{\textbf{Yujie~Lu$^{\dagger}$, Long~Wan$^{\dagger}$, Nayu~Ding$^{\dagger}$, Yulong~Wang$^{\dagger}$, Shuhan~Shen$^{\ddagger,\P}$, Shen~Cai$^{\dagger,\ast}$, Lin~Gao$^{\S,\P,\ast}$}
\thanks{This work was supported in part by the National Natural Science Foundation of China (Grants 62322210, 62206046 and U2033218), Natural Science Foundation of Shanghai (Grant 22ZR1400200), and Fundamental Research Funds for the Central Universities (No. 2232023Y-01).}\\
{$^{\dagger}$Visual and Geometric Perception Lab, Donghua University}\\
{$^{\ddagger}$Institute of Automation, Chinese Academy of Sciences}\\
{$^{\S}$Institute of Computing Technology, Chinese Academy of Sciences}\\
{$^{\P}$University of Chinese Academy of Sciences}\\
{\tt$^{\ast}$Corresponding author: hammer\_cai@163.com, gaolin@ict.ac.cn}\\
{\url{https://github.com/cscvlab/UODFs}}\vspace{-4mm}}

\maketitle

\begin{abstract}\vspace{-3mm}
Neural implicit representation of geometric shapes has witnessed considerable advancements in recent years. However, common distance field based implicit representations, specifically signed distance field (SDF) for watertight shapes or unsigned distance field (UDF) for arbitrary shapes, routinely suffer from degradation of reconstruction accuracy when converting to explicit surface points and meshes. In this paper, we introduce a novel neural implicit representation based on unsigned orthogonal distance fields (UODFs). In UODFs, the minimal unsigned distance from any spatial point to the shape surface is defined solely in one orthogonal direction, contrasting with the multi-directional determination made by SDF and UDF. Consequently, every point in the 3D UODFs can directly access its closest surface points along three orthogonal directions. This distinctive feature leverages the accurate reconstruction of surface points without interpolation errors. We verify the effectiveness of UODFs through a range of reconstruction examples, extending from simple watertight or non-watertight shapes to complex shapes that include hollows, internal or assembling structures.\vspace{-4mm}
\end{abstract}

%------------------%
\section{Introduction}
\label{sec:intro}
\vspace{-2mm}

Neural implicit representation (NIR) of three-dimensional (3D) shapes has emerged as a noteworthy area of research in computer vision and graphics. Currently, the most prevalent NIR for 3D shapes is based on signed distance field (SDF), utilized in various applications such as shape reconstruction~\cite{deepsdf,SIREN,FFN,NI,deepLs,octField,nglod,instant}, or new view synthesis~\cite{NeuS,SparseNeuS,HF-NeuS,VOXURF,MonoSDF}. For any 3D shape, the SDF value of each spatial point reveals two properties: (1) whether the point is inside the shape (indicated by the SDF sign); (2) the minimal distance among all directions from this point to the shape surface (depicted by its absolute value). Despite the advantages offered by SDF based NIR in multiple contexts, there still exist several scenarios where SDF is not applicable.
\begin{figure*}
\centering
  \includegraphics[width=0.95\textwidth]{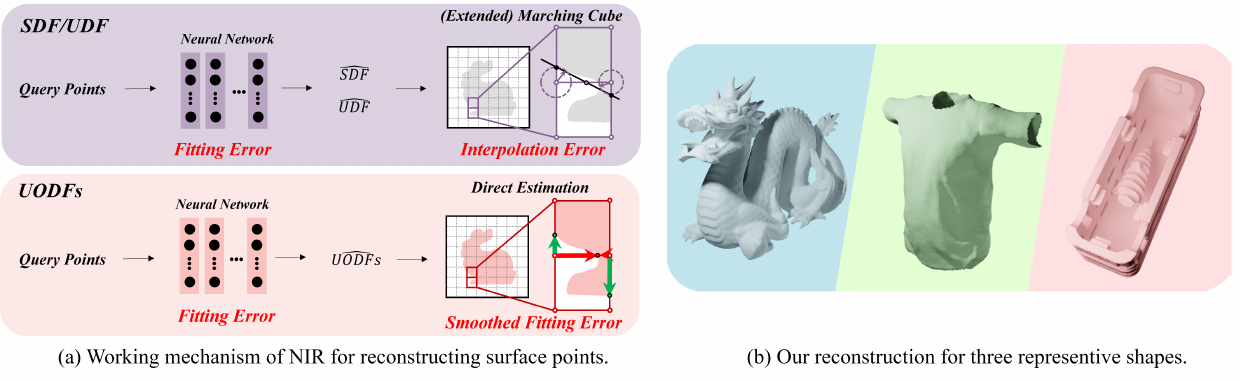}
  \vspace{-4mm}
  \caption{Overview of unsigned orthogonal distance fields (UODFs) based NIR. 
   Unlike SDF or UDF based NIR, which suffers from interpolation errors in surface point reconstruction, UODFs directly estimate surface points and mitigate fitting errors by averaging predictions for each GEP. The upper zoom-in grids illustrate the inaccuracies in traditional methods, where the middle grid edge point (colored purple) is approximately estimated with the distance values (denoted by two dotted circles) of two grid corners. This estimated GEP is far from its true position (colored red) which is jointly predicted by values of our UODFs (two red arrows), as shown in the lower zoom-in grids.
  }
  \label{fig:intro}
  \vspace{-4mm}
\end{figure*}

The first scenario arises when 3D shapes are not watertight or contain internal structures. In such instances, assigning an SDF sign to 3D points might be challenging or inaccurate. To address this limitation and cater to different types of 3D shapes, researchers have proposed the unsigned distance field (UDF) based NIR~\cite{NDF, GIFS, HSDF, meshUDF, 3PSDF, NeUDF}. Losing the sign, a point's UDF value represents the minimal distance traversing all directions to the shape surface, irrespective of the point being inside or outside the model.

The second scenario to consider is the accurate reconstruction of surface points from the NIR of a 3D shape. A common approach for reconstructing surface points from SDF or UDF is the marching cubes (MC) algorithm~\cite{Marching_cubes}. This process is illustrated in the upper part in Fig.~\ref{fig:intro} (a). A neural network makes a prediction of the SDF or UDF values at grid (cube) corners, affected by fitting errors. The MC algorithm applicable to SDF then estimates grid edge points (GEP) through linear interpolation (see the zoom-in grids for detail). However, this approximation can be inaccurate for non-linear local shapes due to interpolation errors. As a result, while employing SDF based NIR, the reconstruction of surface points suffers from these two types of errors.
Similar fitting and interpolation issues are also encountered in deep UDF methods, such as those detailed in~\cite{HSDF,meshUDF,3PSDF}, where the MC algorithm is extended in various ways to incorporate the sign in the UDF at grid corners. Furthermore, in downstream tasks like 3D object classification and part segmentation (see~\cite{PointNet, PointNet++, dgcnn, pct}), shapes are often represented by $1024$ surface points. The implicit representation and reconstruction of surface points at this scale requires large grids in the MC algorithm, further amplifying the interpolation error.

In this paper, we propose the unsigned orthogonal distance fields (UODFs) based NIR. Compared with conventional SDF and UDF, the most distinguishable feature of UODFs is that the minimal distances to the shape surface are defined along three orthogonal directions in 3D space. This means every point in UODFs can directly access its closest surface points along three orthogonal directions, when they exist. Fig.~\ref{fig:intro} (a) depicts how SDF (or UDF) and UODFs based NIR work in surface points reconstruction. All predicted distance values denoted by $\begin{footnotesize}\widehat{...D\!F(s)}\end{footnotesize}$ at grid corners are affected by the fitting error of neural network. However, in contrast to SDF or UDF, the estimation of GEP for UODFs does not introduce interpolation errors. Instead, the fitting error can be mitigated through averaging multiple predictions for each GEP. Therefore, UODFs based NIR naturally leverages accurate reconstruction for diverse 3D shapes, three of which are shown in Fig.~\ref{fig:intro} (b).

The key contributions of this paper are as follows:
\vspace{-2mm}
\begin{itemize}
\item %[1)] 
UODFs based NIR is proposed, which allows for the representation and reconstruction of diverse 3D shapes (such as watertight, non-watertight, multi-layer, or assembling models) in a unified manner. 
\vspace{-2mm}
\item UODFs based NIR enables the direct estimation of surface points from multiple distant sample points along three orthogonal directions, facilitating the interpolation-free reconstruction of grid edge points. 
\vspace{-6mm}
\item UODFs based NIR significantly outperforms traditional SDF or UDF based NIR in terms of reconstruction accuracy, especially for open shapes or when reconstructing small point clouds.
\end{itemize}

%------------------%
\section{Related Works}
\label{sec:relatedWork}
\vspace{-2mm}
The focus in this paper lies in the neural distance field representation of 3D shapes, which could be broadly grouped into three categories for review: SDF based NIR, UDF based NIR, and other types of neural distance field.

%%%%%%%
\vspace{-1mm}
\subsection{SDF based Neural Implicit Representation}
\vspace{-2mm}
Signed distance field (SDF), which implicitly represents a 3D shape as the zero level-set of spatial positions, is popular in the domain of neural implicit representation. A number of methods, including but not limited to DeepSDF~\cite{deepsdf}, IGR~\cite{IGR}, SAL~\cite{SAL}, FFN~\cite{FFN}, SIREN~\cite{SIREN}, NI~\cite{NI}, and GC~\cite{GC_CVPR23}, generally use a multi-layer perceptron (MLP) network to globally fit SDF of 3D shapes. Another kind of methods adopts a local fitting strategy aimed at learning finer shape details, such as OctField~\cite{octField}, DeepLS~\cite{deepLs}, NGLOD~\cite{nglod} and Instant-NGP~\cite{instant}. The marching cubes (MC) algorithm is conventionally employed to extract the zero isosurface of the SDF, enabling simultaneous computation of grid edge points and meshing. In addition, SDF is also adopted to implicitly represent geometric shapes in the task of new view synthesis, such as NeuS~\cite{NeuS}, MonoSDF~\cite{MonoSDF}, and VOXURF~\cite{VOXURF}. These SDF based methods come with the limitation that they can only represent models that are watertight.

\begin{figure*}[th]
  \centering
  \includegraphics[width=0.95\linewidth]{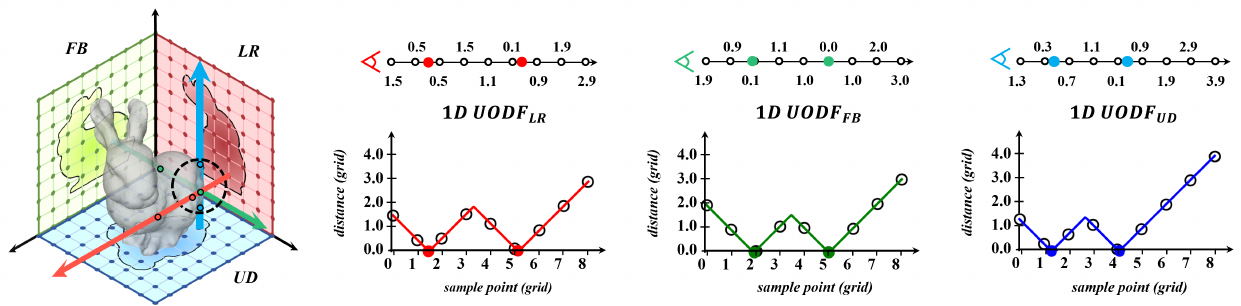}
  \vspace{-4mm}
  \caption{Sketch of UODFs. For understanding the three orthogonal components, 1D derivative, and possible discontinuity between adjacent rays of UODFs, refer to the three characteristics concluded in Sec.~\ref{sec:method_character}.}
  \label{fig:UODFs_character}
 \vspace{-4mm}
\end{figure*}

%%%%%%%
\vspace{-1mm}
\subsection{UDF based Neural Implicit Representation}
\vspace{-2mm}
Unsigned distance field (UDF) based neural implicit methods are proposed to represent arbitrary shapes, such as NDF~\cite{NDF}, CSP~\cite{CSP}, GIFS~\cite{GIFS}, DeepCurrents~\cite{deepcurrents}, 3PSDF~\cite{3PSDF}, NDC~\cite{NDC}, MeshUDF~\cite{meshUDF}, and HSDF~\cite{HSDF}. Within the context of new view synthesis, there also exist works to learn UDF from multi-view images, such as NeUDF~\cite{NeUDF} and Neural UDF~\cite{NeuralUDF}. Although all these approaches represent open shapes, the prediction and meshing processes possess unique distinctions. For instance, NDF firstly obtains discrete points from the predicted UDF, and then employs the ball-pivoting algorithm~\cite{BPA} (BPA) to accomplish meshing; NeUDF follows a similar
process but uses SPSR~\cite{SPSR} for meshing surface points. MeshUDF classifies the signs of grid corners utilizing the UDF gradient, unlike HSDF that predicts the signs and UDF values of grid corners simultaneously. Both methods extend MC~\cite{Marching_cubes} for reconstructing surface points and meshing from UDF, and thus suffer from the fitting error of neural network and the interpolation error of adjacent corners, as shown in Fig.~\ref{fig:intro}.

%%%%%%%
\vspace{-1mm}
\subsection{Other Types of Distance Field}
\vspace{-2mm}
A traditional distance field representation, named directed distance fields (DDFs)~\cite{DDF_siggraph}, was proposed in 2001 to better extract surface points on sharp edges. DDFs are calculated explicitly and used for subsequent contour reconstruction by extending the MC algorithm. Although our proposed UODFs based NIR relates to DDFs in terms of distance field definition, the focal points of our study involve fitting UODFs using neural networks and reconstructing GEP, which distinguishes our work from~\cite{DDF_siggraph}. 

There exist some works~\cite{DDF_CVPR22, NeuralODF} studying deep distance fields along arbitrary directions for 3D shape representation. While these approaches exhibit flexibility in reconstructing surface points, the accuracy does not match the levels achieved by SDF or UDF based NIR.

\vspace{-2mm}
%------------------%
\section{Method}
\vspace{-2mm}

%%%%%%%
\subsection{Definition of UODFs}
\label{sec:method_definition}
\vspace{-2mm}
The unsigned orthogonal distance of a point is defined as the closest distance from this point to the shape surface, along one orthogonal direction. The term `unsigned' means the distance value can never be negative, and there is no distinction between the inside and outside of the shape. In 3D space, there are three orthogonal directions, denoted by `left-right (LR)', `front-back (FB)' and `up-down (UD)' respectively. Therefore, UODFs portray each shape as a combination of three distance fields along these orthogonal directions, which are denoted by $U\!O\!D\!F_{L\!R}$, $U\!O\!D\!F_{F\!B}$ and $U\!O\!D\!F_{U\!D}$, respectively.
The ground truth values of UODFs can be computed by ray stabbing along the three orthogonal directions.
For a normalized model $\mathcal{A}$, its surface is denoted by $\mathcal{S}$. UODFs are limited in the normalized cube space $\mathcal{R}$ ($\mathcal{R}\!\in\![-1,1]^3$). 
Denote the ray crossing a point $p$ along one orthogonal direction $X$ by $ray(p,X)$ ($X\!\in\!\{L\!R, F\!B, U\!D\}$). 
The subset $D(p,X)$ denotes all intersections of the surface $\mathcal{S}$ with $ray(p,X)$. UODFs of $p$ is expressed by 
\vspace{-6mm}

\begin{equation}
\small
    U\!O\!D\!Fs(p) = \{ U\!O\!D\!F_{L\!R}(p), U\!O\!D\!F_{F\!B}(p), U\!O\!D\!F_{U\!D}(p) \}
\end{equation}

\vspace{-2mm}
\noindent where each UODF of the point is defined by
\vspace{-6mm}

\begin{equation}
\small
U\!O\!D\!F_X(p) = \underset{q \in D(p,X)}{\arg\min}(|p - q|), 
\quad \mathrm{if} \,\, D(p,X)\!\neq\!\varnothing
\label{equ:2}
\end{equation}

\vspace{-3mm}
If $D(p,X)$ is empty, indicating $ray(p,X)$ does not intersect with $\mathcal{S}$, $U\!O\!D\!F_X(p)$ is not defined by a specific value.

\begin{figure}[t]
  \centering
  \includegraphics[width=0.9\linewidth]{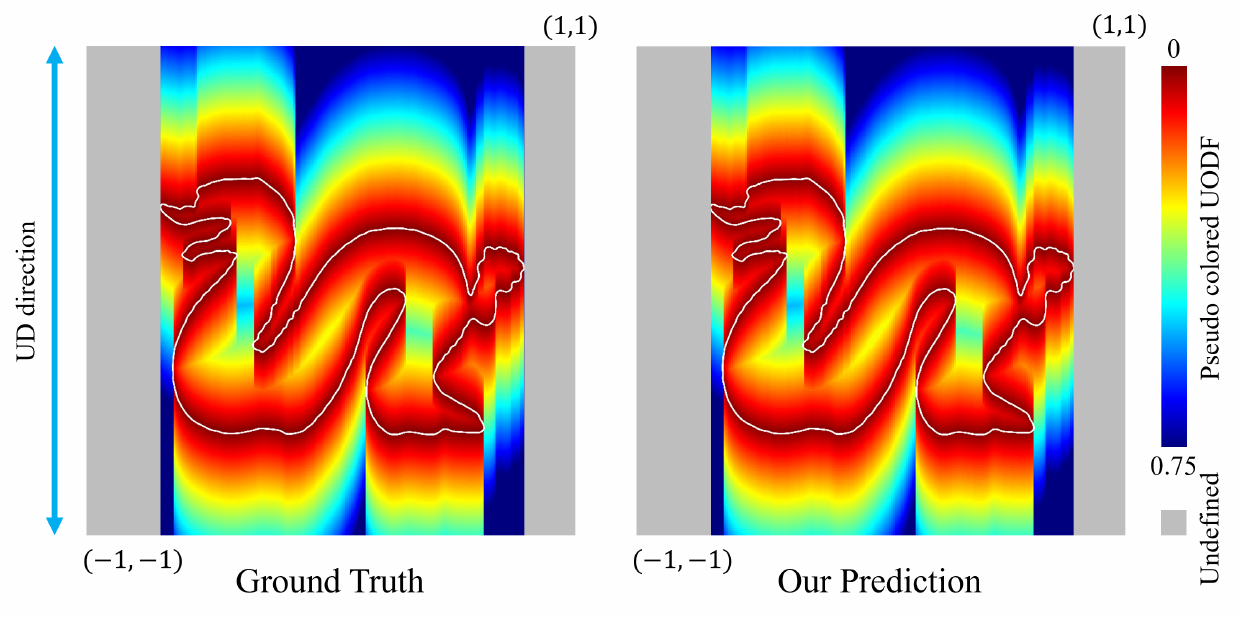}
  \vspace{-5mm}
  \caption{One slice from the $U\!O\!D\!F_{U\!D}$ of the 3D model `Dragon'. Discontinuity between adjacent ‘UD’ rays occurs when there is a change in intersection with the surface. The grayscale areas denote undefined $U\!O\!D\!F_{U\!D}$ in ground truth calculation and our prediction.}
  \label{fig:2Dfitting_Dragon}
 \vspace{-12mm}
\end{figure}

%%%%%%%
\vspace{-2mm}
\subsection{Characteristics of UODFs}
\label{sec:method_character}
\vspace{-2mm}

UODFs possess several characteristics that significantly distinguish them from SDF or UDF.
\begin{figure*}[t]
  \centering
    \includegraphics[width=0.99\linewidth]{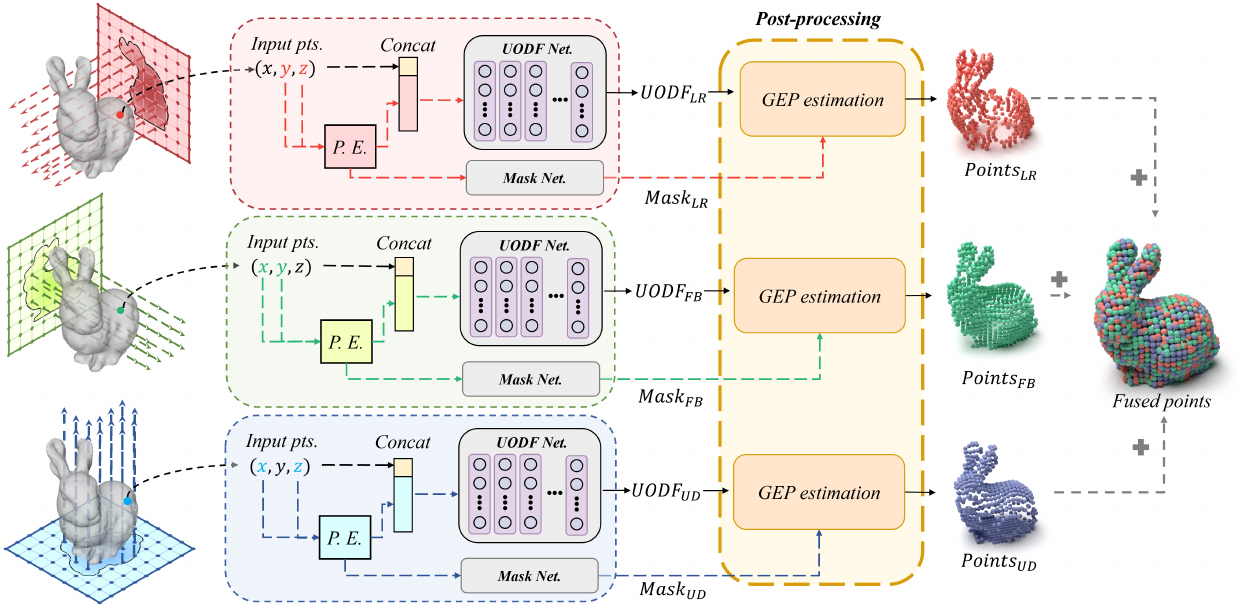}
    \vspace{-2mm}
  \caption{Network architecture and processing pipeline. Each UODF is individually regressed using a UODF and mask network. The grid edge points in one orthogonal direction can be estimated from its UODF, followed by points fusion to form final reconstruction results.}
  \label{fig:architecture}
  \vspace{-5mm}
\end{figure*}

\textit{Characteristic 1}. UODFs consist of three unsigned distance fields along orthogonal directions. Any ray in an orthogonal direction corresponds to a 1D unsigned distance field. Thus, $U\!O\!D\!F_X$ can be seen as a collection of 1D unsigned distance fields for all parallel rays in the orthogonal direction $X$. Fig.~\ref{fig:UODFs_character} depicts three orthogonal rays (red, green, and blue) of a point, their intersections with the shape (two surface points for each ray), and their 1D unsigned distance fields.
Here rays and sample points (hollow circles) locate at grid corners, which will be explained in detail in Sec.~\ref{sec:method_Sampling} and Sec.~\ref{sec:method_pointsEstimation}.

\textit{Characteristic 2}. The absolute value of 1D derivative of each UODF for each point is equal to $1$. Fig.~\ref{fig:UODFs_character} also demonstrates that the UODF values of equally-spaced points on one side
of the surface point form an arithmetic sequence. Each line segment in the 1D UODF diagrams is at an angle of $\pm45$ degrees. This characteristic can be utilized to mitigate estimation errors of surface points from UODF predictions of multiple sample points.

\textit{Characteristic 3}. $U\!O\!D\!F_{X}$ may display discontinuity between adjacent parallel rays, which distinguishes it from the SDF or UDF that is continuous everywhere. Fig.~\ref{fig:2Dfitting_Dragon} depicts a slice of $U\!O\!D\!F_{U\!D}$ from the 3D model `Dragon'. It is evident that the $U\!O\!D\!F_{U\!D}$ values of two neighborly points on adjacent rays can differ greatly. This unique characteristic mirrors the
behavior observed in laser sensors and facilitates the direct and accurate prediction of fine details.

In summary, these three characteristics of UODFs are beneficial to the accurate estimation of surface points, especially for diverse complex shapes or small point clouds. Since each UODF is defined on parallel rays, the UODFs in the non-occluded outside regions are basically equivalent to the depth maps captured from six facets of a unit cube. However, UODFs additionally manage the occluded outside and inside regions of 3D models in a unified manner. 

\vspace{-1mm}
%%%%%%%
\subsection{Network Architecture and Processing Pipeline}
\label{sec:method_pipeline}
\vspace{-2mm}

Fig.~\ref{fig:architecture} depicts the proposed network architecture and processing pipeline. Parallel rays are defined on each orthogonal plane. For a spatial point, its 2D coordinates on one orthogonal plane are employed to extract a $42$-dimensional feature via position encoding (`P. E.' module). This ray feature, combined with the 3D coordinates, forms the $45$-dimensional input for a multi-layer perceptron (MLP) network to regress the UODF in this orthogonal direction. Meanwhile, to avoid regressing the UODF of non-intersected rays which are not defined in Eq.~\ref{equ:2}, the ray feature is utilized to regress a 2D mask on this orthogonal plane (see the area in each silhouette). Subsequently, the surface points in this orthogonal direction can be estimated from rays that intersect the shape, which is illustrated in Sec.~\ref{sec:method_pointsEstimation}.

The networks to regress the UODF and the mask in each orthogonal direction are a $10$$*$$256$ MLP and $3$$*$$256$ MLP, respectively. The supervised mask values are set to be `0' and `1', denoting the outside and inside of the silhouette respectively. The loss function of $U\!O\!D\!F_{X}$ is composed of three component loss terms:
\vspace{-3mm}

\begin{equation}
\small
\mathcal{L}_{all} = \lambda_1*\mathcal{L}_{value} + \lambda_2*\mathcal{L}_{der} + \lambda_3*\mathcal{L}_{pred}.
\label{equ:allLoss}
\end{equation}

\vspace{-1mm}
\noindent where the weights are experimentally set to: $\small \lambda_1\!=\!3000$, $\small \lambda_2\!=\!50$, and $\small \lambda_3\!=\!1000$.

For any point $p$, the term $\mathcal{L}_{value}$ means that its predicted UODF value \begin{small}
$\widehat{U\!O\!D\!F_X}(p)$\end{small} should be equal to the ground truth $U\!O\!D\!F_X(p)$, which is formulated as 
\vspace{-3mm}

\begin{equation}
\small
\mathcal{L}_{value} = |\,\, \widehat{U\!O\!D\!F_X}(p)-U\!O\!D\!F_X(p) \,\,|.
\end{equation}

\vspace{-1mm}
According to `\textit{Characteristic 2}' analyzed in Sec.~\ref{sec:method_character}, the term $\mathcal{L}_{der}$ denotes the absolute value of UODF derivative for any point should be equal to $1$, which is expressed by 
\vspace{-3mm}

\begin{equation}
\small
\mathcal{L}_{der} = |\,\, |U\!O\!D\!F_X^{'}(p)|-1 \,\,|.
\end{equation}

\vspace{-1mm}
Inspired by the recent works~\cite{HSDF, GC_CVPR23}, the term $\mathcal{L}_{pred}$ is extra added to constrain the UODF value of the predicted surface point of $p$, which is expressed by 
\vspace{-6mm}

\begin{equation}
\small
\mathcal{L}_{pred} = |\,\, U\!O\!D\!F_X(\, p - sign(U\!O\!D\!F_X^{'}(p))*U\!O\!D\!F_X(p) \,) \,\,|.
\end{equation}

\vspace{-2mm}
\noindent where \begin{small}$sign(U\!O\!D\!F_X^{'}(p))$\end{small} denotes the direction of derivative. Consequently, \begin{small}$p\!-\!sign(U\!O\!D\!F_X^{'}(p))\!*\!U\!O\!D\!F_X(p)$\end{small} represents the predicted surface point based on $p$ and $U\!O\!D\!F_X$.

\vspace{-1mm}
%%%%%%%
\subsection{Sampling During Inference and Training}
\label{sec:method_Sampling}
\vspace{-2mm}

The sampling process entails two steps: rays sampling upon orthogonal planes and points sampling along rays. During inference, both rays and points sampling are discretized. For the sake of comparison with MC~\cite{Marching_cubes} and MeshUDF~\cite{meshUDF}, rays and points are sampled at grid corners with equal intervals in our experiments. However, UODFs also facilitate an arbitrary 2D and 1D pattern for rays and points sampling, respectively.

During training, due to the possible discontinuity between adjacent rays analyzed as `\textit{Characteristic 3}' in Sec.~\ref{sec:method_character}, rays sampling is retained as discrete. We densely sample each ray on three orthogonal planes with a resolution $257$*$257$. For points sampling along rays, we uniformly sample $256$ points per 10 epochs. This accommodates the continuous nature of 1D UODF along each ray, allowing for any given point on the ray to accurately predict its closest surface point.

%%%%%%%
\vspace{-1mm}
\subsection{Surface Points Estimation for Each Ray}
\label{sec:method_pointsEstimation}
\vspace{-1mm}

\begin{figure}[t]
  \centering
\includegraphics[width=0.75\linewidth]{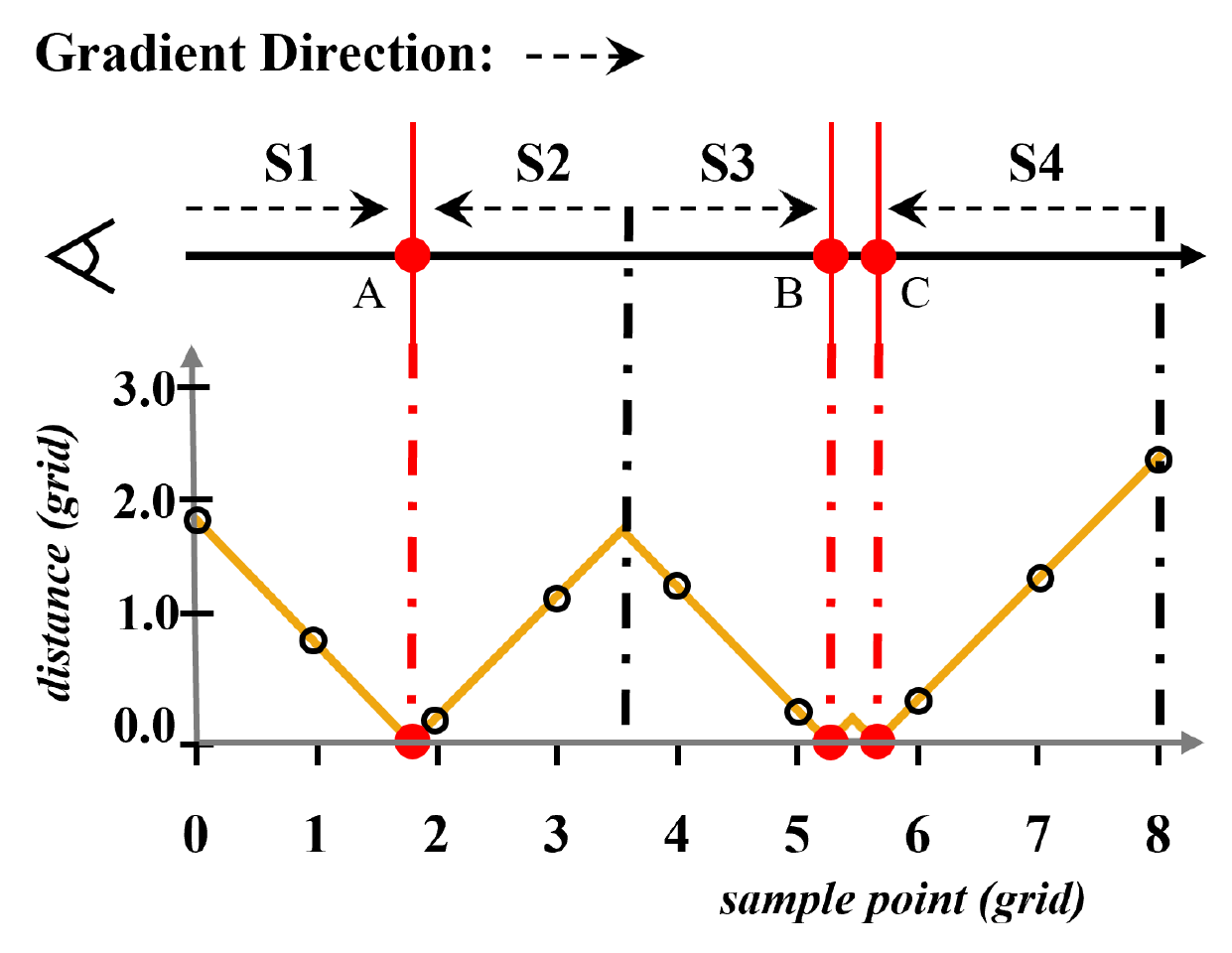}
  \vspace{-5mm}
  \caption{Surface points estimation along a ray.}
  \label{fig:estimationPoints}
  \vspace{-6mm}
\end{figure}

Although points sampling along rays in inference could be random for UODFs, we sample points with equal intervals in alignment with MC's resolution. Subsequently, surface points along each ray are estimated from these equally spaced sample points. Fig.~\ref{fig:estimationPoints} illustrates how the surface points along a ray are estimated. Three intersection points, labelled `A', `B', and `C', are represented by solid circles along this ray. The lower part of the figure depicts the 1D UODF along this ray, with nine equally spaced points (hollow) sampled. Unlike SDF or UDF, each sample point directly predicts its nearest surface point. 
As a result, our UODFs based NIR can potentially estimate a surface point from several sample points exhibiting consistent gradients. For instance, the nine sample points on this ray are divided into four segments, each of which estimates one surface point by averaging the predictions of all sample points within this segment. Both segments `S1' and `S2' estimate a surface point `A'. If the distance between the two estimated points is smaller than a threshold $\tau$ (with $\tau$=$1/512$ in our experiments), these two estimated points will be averagely merged into a new one.  
For segments `S3' and `S4', if the distance between their estimation points is greater than $\tau$, both the surface points `B' and `C' are reconstructed, although they are on the same grid edge. 
This phenomenon occurs when two closed models are close together or a thin plate model is represented by two planes. However, the MC algorithm is limited to estimating at most one intersection point on each grid edge, accompanying the introduction of the interpolation error. In contrast, our UODF based estimation can reconstruct two intersection points on one gird edge and smooth out the predictions of multiple sample points, influenced by the fitting error of the network.

Although the surface point estimation in UODFs makes use of distance values from sample points on a ray, it is quite different from another point estimation technique suitable to SDF and UDF known as sphere tracing~\cite{sphereTracing}. Sphere tracing adopts a conservative strategy, only permitting ray stepping in accordance with the unsigned distance from the current position. On the contrary, one point in UODFs directly predicts the closest point in an orthogonal direction, irrelevant to the ray tracing scheme. 

After estimating surface points in three orthogonal edges, a fusion operation is engaged to derive the final GEP, as shown in Fig.~\ref{fig:architecture}. Specifically, if there are estimated points on at least 3 adjacent edges, those points are retained. Otherwise, they will be deleted. This strategy is similar to MC and allows for the effective filtering of isolated points.

\vspace{-1mm}
%%%%%%%
\subsection{Mesh Extraction}
\label{sec:method_meshRecovery}
\vspace{-1mm}

In the aforementioned processing pipeline, UODFs based NIR reconstructs surface points for arbitrary shapes in a unified manner. If required, an additional meshing algorithm can be applied to extract mesh from surface points. We apply a masked Poisson method for mesh extraction from UODFs, as recommended by NeUDF~\cite{NeDDF}. Specifically, the screened Poisson surface reconstruction (SPSR) technique~\cite{SPSR} is used to extract a watertight mesh, followed by masking out the spurious triangles that have a considerable distance from the reconstructed points.

Despite the SPSR method, without the fine-tuning of the default hyper-parameters, generates superior mesh results in our experiments, we explore another way for simultaneous GEP reconstruction and mesh extraction, based on the MeshUDF~\cite{meshUDF} algorithm. These findings are preliminary and will be refined in subsequent research.

% To summarise, we compute the normal vectors and UDF values at grid corners from our predicted UODFs and feed them into MeshUDF as an input. Detailed process and experimental results are available in the supplementary material.

\vspace{-1mm}
%------------------%
\section{Experiments}
\label{sec:exp}

%%%%%%%
\vspace{-1mm}
\subsection{Dataset and Metrics}
\label{sec:exp_dataset}
\vspace{-1mm}

\noindent \textbf{Dataset.} Over 50 models from a variety of datasets are used to verify the reconstruction accuracy of diverse shapes. Specifically, 36 shapes come from Thingi10K~\cite{thingi10K}, including 32 shapes in the Thingi32 subset, which is used in NI~\cite{NI} and NGLOD~\cite{nglod}.
Other shapes come from the Stanford 3D Scanning Repository~\cite{stanford}, ShapeNet~\cite{shapeNet}, the MGN dataset~\cite{bhatnagar2019mgn}, and our own self-generation. 

\vspace{1mm}

\noindent \textbf{Metrics.} The reconstruction accuracy of grid edge points (GEP) is evaluated by a $\textit{L}_2$ Chamfer distance (CD), abbreviated as `GEP-CD'.
For mesh, a $\textit{L}_2$ CD and normal consistency (NC) of $100,000$ evenly distributed points are measured, abbreviated as `Mesh-CD' and `Mesh-NC' respectively. 
For the sake of fair comparison, all reconstructed meshes from tested NIR methods employ predicted distance values of $257^3$ grid corners, except the experiments with multiple grid resolutions as shown in Sec.~\ref{sec:exp_resolution}.

%%%%%%%
\vspace{-1mm}
\subsection{Reconstruction of Watertight Shapes}
\label{sec:exp_watertight}
\vspace{-1mm}
The first experiment is to verify the reconstruction accuracy of watertight shapes, of which the SDF sign is typically computed with high accuracy.

Table~\ref{tab:thingi31} exhibits the reconstruction metrics on the Thingi32 dataset, with comparative baselines established by SIREN~\cite{SIREN} and NGLOD~\cite{nglod}. For SIREN, the normal information is not utilized to make the same supervision signal across all methods.
For NGLOD, we test two levels of detail: LOD3 and LOD5, each with prior octree knowledge at maximum resolutions of $16^3$ and $64^3$, respectively. 
In both CD metrics, UODFs outperforms the SOTA NGLOD5, even though it has the advantage of locally fitting SDF.
From the table, it is also evident that the meshing technique significantly amplifies the CD, resulting in a decrease in reconstruction accuracy, whether using the MC algorithm used by NGLOD or the masking Poisson method employed in our work.

\begin{figure*}[t]
  \centering
    \includegraphics[width=0.97\linewidth]{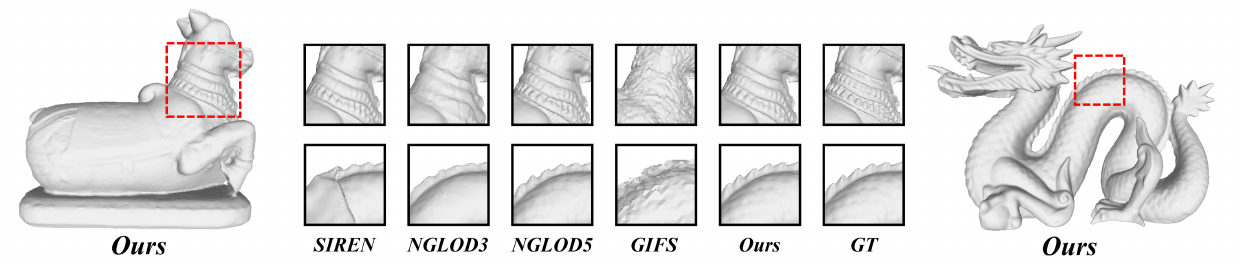}
    \vspace{-4mm}
  \caption{Reconstructed meshes for two watertight shapes.}
  \label{fig:singleWatertight}
  \vspace{-4mm}
\end{figure*}

Fig.~\ref{fig:singleWatertight} visualizes the reconstructed meshes for one Thingi32 shape `‘Nandi the Bull' and one Stanford Scanned shape `Dragon'.
Except the three SDF based methods, the NIR method GIFS~\cite{GIFS} dealing with arbitrary shapes also participates in the comparison. The zoom-in figures indicate that our UODFs based NIR can accurately reconstruct the fine details of these watertight shapes, but the UDF based GIFS fails.

%%%%%%%
\vspace{-1mm}
\subsection{Reconstruction of Non-watertight Shapes}
\label{sec:exp_nonWatertight}
\vspace{-1mm}
In this subsection, we evaluate the reconstruction performance on non-watertight shapes. The comparison involves three UDF based methods, including NDF~\cite{NDF}, HSDF~\cite{HSDF}, GIFS~\cite{GIFS}. For the training of the UDF of individual shapes, $2$M spatial points are sampled. Table~\ref{tab:MGN10} shows the metrics of $10$ representative garments from the MGN dataset. The performance of the proposed UODFs based method substantially surpasses the other UDF based methods. It is worth noting that the accuracy metrics of the non-watertight models reconstructed by us are on the same level as those of the watertight models shown in Table~\ref{tab:thingi31}. However, compared with the SOTA SDF based NGLOD, UDF based methods generally perform unsatisfactorily, due to different fitting strategies and the introduction of additional errors.
For example, HSDF utilizes the shape feature extracted by 3D CNN to globally fit UDF and additionally requires to classify the inside and outside of grid corners.

\begin{table}[t]
\begin{center}
\caption{Average reconstruction accuracy on Thingi32 dataset.}
\label{tab:thingi31}
\vspace{-3mm}
\resizebox{0.48\textwidth}{!}{
\begin{tabular}{|l||c|c|c||c|}
\hline
\multirow{2}{*}{\diagbox{Metric}{Method}} & \multicolumn{3}{c||}{SDF} & UODFs \\
\cline{2-5}
& \tabincell{c}{SIREN} & \tabincell{c}{NGLOD3} & \tabincell{c}{NGLOD5}  & \textbf{Ours} \\ 
\hline
CD-GEP (*$10^5$) ~$\bm{\downarrow}$ & 149 & 0.684 & 0.432 & \textbf{0.378} \\
\hline 
CD-Mesh (*$10^5$)~$\bm{\downarrow}$ & 147 & 2.99 & 2.83 & \textbf{2.69}
\\
\hline
NC-Mesh ~$\bm{\uparrow}$ & 92.3 & 98.0 & \textbf{98.5} & 98.4 \\
\hline
\end{tabular}
}
\end{center}
\vspace{-6mm}
\end{table}
\begin{table}[t]
\begin{center}
\caption{Average reconstruction accuracy on MGN10 dataset.}
\label{tab:MGN10}
\vspace{-3mm}
\resizebox{0.43\textwidth}{!}{
\begin{tabular}{|l||c|c|c||c|}
\hline
\multirow{2}{*}{\diagbox{Metric}{Method}} & \multicolumn{3}{c||}{UDF} & UODFs \\
\cline{2-5}
& \tabincell{c}{NDF} & \tabincell{c}{HSDF} & \tabincell{c}{GIFS}   & \textbf{Ours} \\ 
\hline
CD-GEP (*$10^5$) ~$\bm{\downarrow}$ & 128 & 13.0 & 4.95  & \textbf{0.227} \\
\hline 
CD-Mesh (*$10^5$) ~$\bm{\downarrow}$ & 128 & 14.4 & 6.1  &\textbf{1.93} %\textcolor{red}{\textbf{3.19}}
\\
\hline
NC-Mesh~$\bm{\uparrow}$ & 90.8 & 88.0 & 92.0  &\textbf{99.6}\\
\hline
\end{tabular}
}
\end{center}
\vspace{-6mm}
\end{table}

\begin{figure*}[th]
  \centering
    \includegraphics[width=0.96\linewidth]{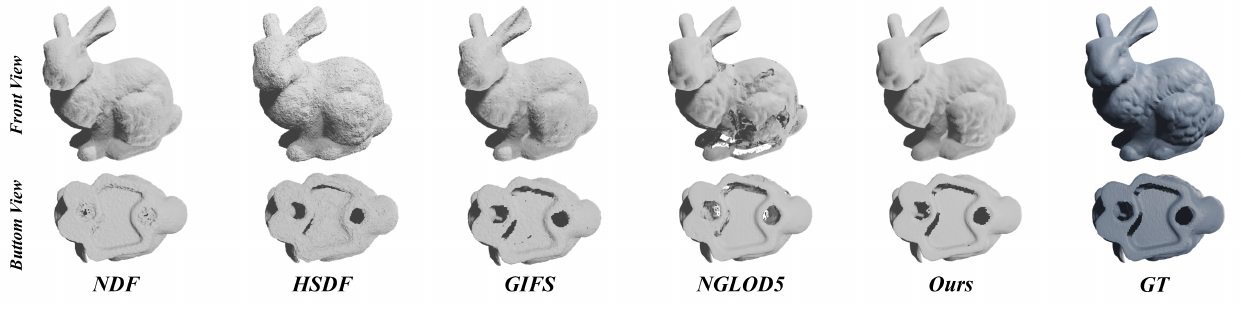}
    \vspace{-4mm}
  \caption{Reconstructed meshes for the non-watertight `Bunny'.}
  \label{fig:singleNonWatertight}
  \vspace{-4mm}
\end{figure*}

As displayed in Fig.~\ref{fig:singleNonWatertight}, we show the reconstructed meshes for the model ‘Bunny’ from the Stanford dataset, which is unclosed at its bottom. Although the three UDF based methods reconstruct the complete model without significant holes that should not exist, only our method is capable of reconstructing finer details. The NGLOD5 method dealing with watertight models also participates in this comparison. However, as the SDF sign is incorrectly calculated near the bottom of the model, there exist obvious artifacts.

\vspace{-1mm}
%%%%%%%
\subsection{Reconstruction of Complex Shapes}
\vspace{-1mm}
The shapes in the above two subsections are structurally simple even though they have rich surface details. In this subsection, we test various shapes with complex structure.
Reconstruction results of two complex shapes as examples are shown in Fig.~\ref{fig:complex}. The first one is `Hilbert Cube' from the Thingi10K dataset. Although this shape is watertight, the topology is challenging since it contains cavities on a large scale. Moreover, the SDF values may be calculated incorrectly in local areas. 
The second shape is an assembling model we generate, consisting of the internal multi-layer non-watertight fish and the external hollow yet watertight box.
It can be seen from the zoom-in views that the three UDF based methods reconstruct the rough surface of these two complex shapes. NGLOD5 reconstructs the details of the watertight shape, but fails for the internal multi-layer fish. Our UODFs significantly outperforms the others, displaying its superiority.

\vspace{-1mm}
%%%%%%%
\subsection{Ablation of Loss Function}
\label{sec:exp_ablation}
\vspace{-1mm}

\begin{figure*}[t]
  \centering
    \includegraphics[width=\linewidth]{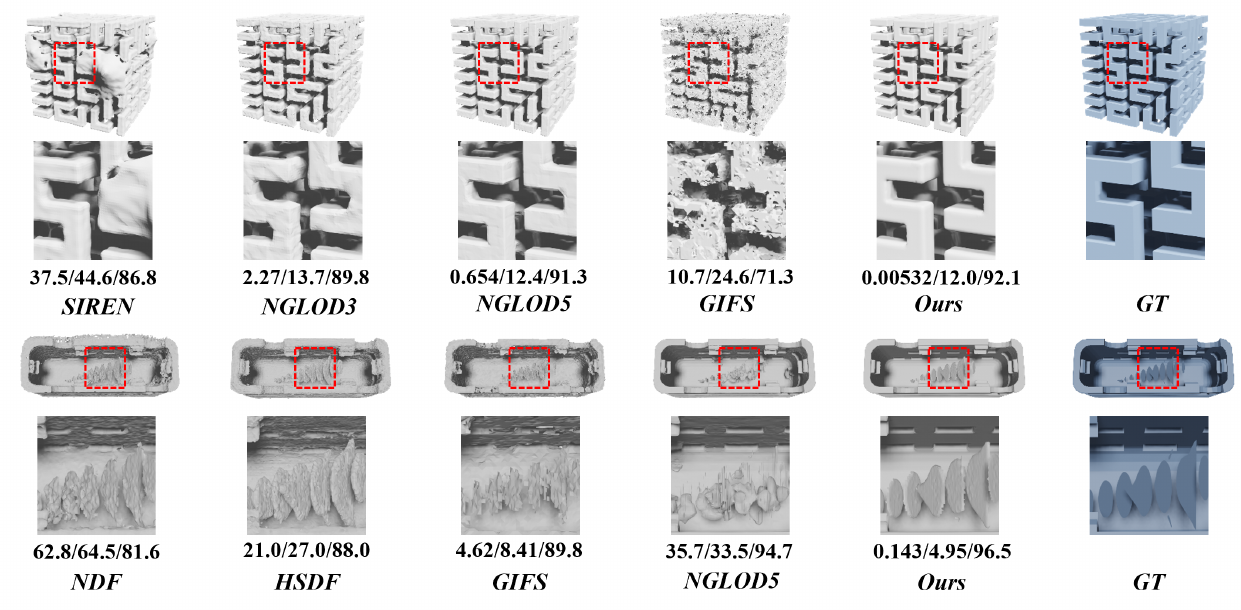}
  \vspace{-8mm}
  \caption{Reconstructed meshes for two representative shapes with complex structure. The metrics from left to right below each shape are CD-GEP(*$10^5$)~$\bm{\downarrow}$, CD-Mesh(*$10^5$)~$\bm{\downarrow}$, and NC-Mesh~$\bm{\uparrow}$, respectively.}
  \label{fig:complex}
  %\vspace{2mm}
\end{figure*}

An ablation study is conducted for the proposed loss function. %, detailed in this subsection. 
Specifically, we use three distinct shapes, which are the watertight `Nandi the Bull' (in Fig.~\ref{fig:singleWatertight}), the non-watertight `Bunny' (in Fig.~\ref{fig:singleNonWatertight}), and the complex `Hilbert Cube' (in Fig.~\ref{fig:complex}), to evaluate the reconstruction accuracy under varying loss configurations. Table~\ref{tab:ablation} reveals that most results differ slightly, mainly because the loss term $\mathcal{L}_{value}$ plays a most important role in our supervised training.

\vspace{-1mm}
%%%%%%%
\subsection{Reconstruction at Multi-resolution Grids}
\label{sec:exp_resolution}
\begin{figure*}[ht]
  \centering
    \includegraphics[width=\linewidth]{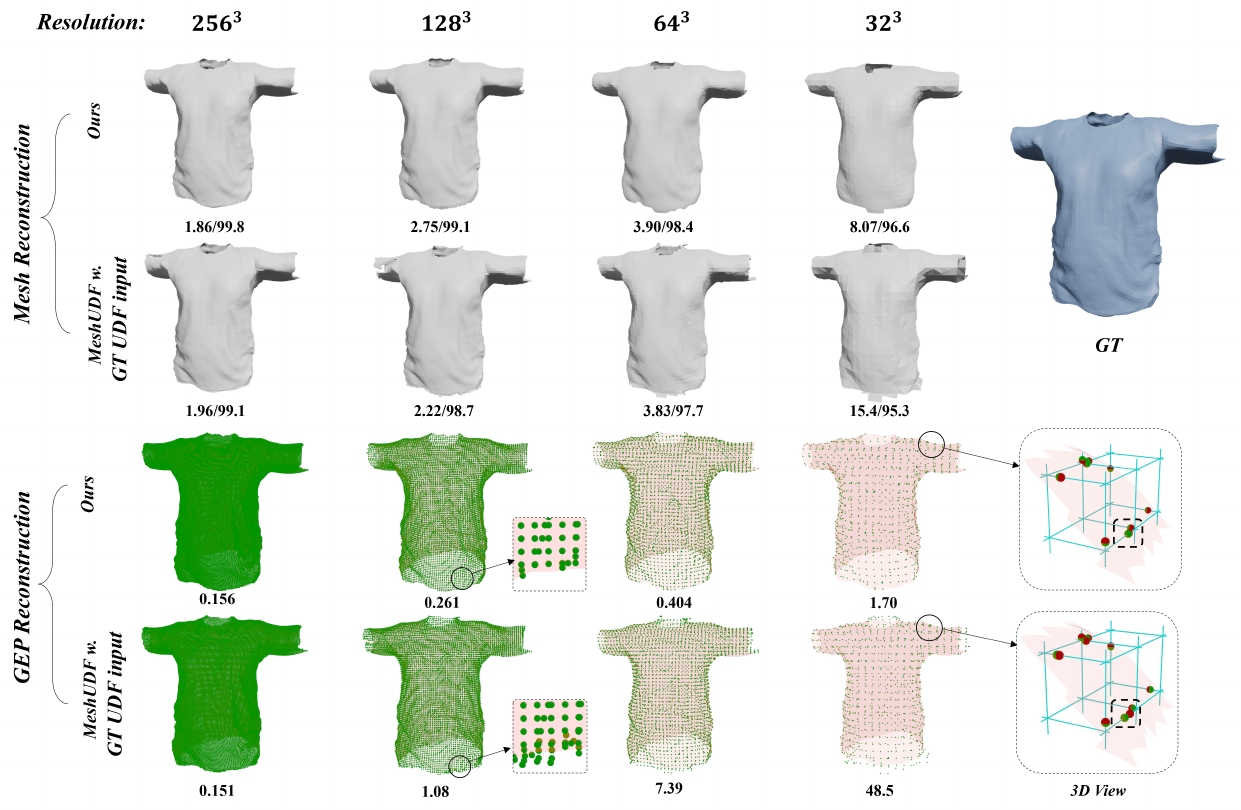}
  \vspace{-8mm}
  \caption{Reconstruction results at various resolutions of grids. The metrics from left to right below each mesh are CD-Mesh(*$10^5$)~$\bm{\downarrow}$ and NC-Mesh~$\bm{\uparrow}$ respectively. The metric below each GEP result (along with red ground truth model) is CD-GEP(*$10^5$)~$\bm{\downarrow}$. The zoom-in views in the middle highlight the spurious GEP results of MeshUDF. The zoom-in 3D views in the right draw local grids and ground truth GEP (red spheres), demonstrating the deviation of GEP reconstruction caused by the interpolation error (highlighted by black dotted boxes).}
  \label{fig:resolution}
\end{figure*}

\vspace{-1mm}
In this subsection, we conduct reconstructions at varying grid resolutions, ranging from $32^3$ to $256^3$. Fig.~\ref{fig:resolution} separately displays the reconstructed meshes and grid edge points (GEP) of a T-shirt. To demonstrate the superior reconstruction accuracy of our UODF-based NIR, we compare it with MeshUDF~\cite{meshUDF}, which is an extended marching cube algorithm designed for UDF. MeshUDF is employed to use ground truth UDF values and gradients at grid corners as input. It is worth noting that in this scenario, the reconstruction results of MeshUDF do not suffer from fitting errors of neural network, but are only affected by interpolation errors. For this non-watertight shape, our method consistently outperforms MeshUDF at all tested grid resolutions, even though MeshUDF uses ground truth UDF as input. Additionally, MeshUDF tends to create spurious meshes around the edge of the clothing, particularly noticeable at lower resolutions.
We present the GEP reconstruction results in the lower part of the figure. The metric values shown, along with the zoomed-in views (located in the middle and right areas), clearly illustrate that our reconstructed GEP are in closer proximity to the model surface.

\begin{table}[t]
\begin{center}
\caption{Ablation study of loss function on three shapes. The metrics in each cell are CD-GEP(*$10^5$)~$\bm{\downarrow}$, CD-Mesh(*$10^5$)~$\bm{\downarrow}$, and NC-Mesh~$\bm{\uparrow}$, respectively. }
\label{tab:ablation}
\vspace{-3mm}
\resizebox{0.48\textwidth}{!}{
\begin{tabular}{|l||c|c|c|}
\hline
\diagbox{Loss}{Shape} & `Nandi the Bull' & `Bunny' & `Hilbert Cube' \\
\hline
All & \small{\textbf{0.228} / \textbf{2.77} / \textbf{98.5}} & \small{0.170 / \textbf{2.73} / \textbf{99.2}} & \small{\textbf{0.00532} / \textbf{12.0} / \textbf{92.1}}\\
% \small{\textbf{0.0399} / \textbf{2.43} / \textbf{98.8}} \\
\hline 
w/o $\mathcal{L}_{der}$ & \small{0.238 / 2.79 / 98.5} & \small{0.161 / 2.93 / 99.1} & \small{0.00663 / 12.0 / 92.0} \\ %\small{0.0509 / 2.43 / 98.8} \\
\hline
w/o $\mathcal{L}_{pred}$ & \small{0.297 / 2.79 / 98.4} &\small{\textbf{0.152} / 2.93 / 99.2} &  \small{0.256 / 12.3 / 91.9}\\
\hline

\end{tabular}
}
\end{center}
\vspace{-8mm}
\end{table}

\vspace{-1mm}
%%%%%%%
\subsection{Discussion and Limitation}
\label{sec:exp_Limitation}
\vspace{-1mm}

Additional experimental results are available in the supplementary material, where UODFs based NIR continues to exhibit exceptional reconstruction performance. This achievement can be attributed to several key factors discussed below:

1) Each UODF enjoys a similar working mechanism to a planar laser, capable of accurately measuring distances to the closest surface points along a direction. Notably, UODF further works within occluded or internal regions.

2) It is verified by us that a discontinuous UODF can be fitted well by an MLP network. This discovery challenges the traditional belief that only continuous fields (e.g., SDF) are suitable for neural network fitting. 

3) Interpolation-free estimation of GEP in one orthogonal direction, as well as the fusion of points across three directions, is robust to any model tested. If a mesh is required, the adopted masked SPSR also performs robustly.

A notable limitation of UODFs is the necessity of three neural networks, each needed to fit one UODF. This requirement leads to a relatively large number of parameters in neural networks at present. While certain correlation has been observed between distance fields in different directions, the prospect of fitting three fields within just one neural network is an area of interest for future exploration.

\vspace{-2mm}
%------------------%
\section{Conclusion}
\vspace{-2mm}
In this paper, we propose unsigned orthogonal distance fields (UODFs) based NIR for accurate reconstruction of diverse 3D shapes. UODFs diverge from conventional SDF and UDF in their unique characteristics, which include combining each UODF across three orthogonal directions, estimating surface points directly from distant sample points, and exhibiting discontinuities between rays. Consequently, specific neural networks are required for UODFs regression, along with post-processing methods for surface points reconstruction. 
Thorough experiments on more than $50$ diverse shapes validate that UODFs consistently outrank all competitors in a unified reconstruction manner. UODFs could even outperform traditional meshing methods that use ground truth SDF or UDF input data. In the future, we plan to explore more applications of UODFs, such as in real-time rendering of shapes. The feasibility of fitting UODFs within a single neural network is also worth pursuing.
\newpage

\clearpage
\newpage
\small
\bibliographystyle{ieee_fullname}
% \bibliography{my}

\normalsize

\twocolumn[
\begin{@twocolumnfalse}
\section*{\centering{Supplementary Material for \\ \emph{Unsigned Orthogonal Distance Fields (UODFs): } \\ \emph{An Accurate Neural Implicit Representation for Diverse 3D Shapes\\[25pt]}}}
\end{@twocolumnfalse}
]

\appendices

\renewcommand{\theequation}{\thesection.\arabic{equation}}

\renewcommand{\thefigure}{\thesection.\arabic{figure}}
\setcounter{figure}{0}

\renewcommand{\thetable}{\thesection.\arabic{table}}
\setcounter{table}{0}

%\section*{Appendix}

%------------------%
\section{Detailed Configurations}

%------------------%
\subsection{Training Details in Our method}
All reconstruction results shown in this manuscript are free of fine-tuning parameters in the network and the post-processing method, thus are easy to reproduce. Each shape is normalized with a radius of $0.9$. In our training, the UODFs for each model are trained for $100$ epochs. The method for point sampling during training is detailed in Sec.~\textcolor{red}{3.4} in the main manuscript. We adopt a batch size of $1024$ points during training. The optimization is performed using the Adam optimizer, starting with an initial learning rate of $0.001$, which is halved every $20$ epochs.

%------------------%
\subsection{Post-processing Details in Our method}

In this section, we detail the post-processing meshing process for UODFs. Our process begins with the employment of the VCGlib library~\cite{VCGlib}, a comprehensive toolbox also utilized by MeshLab~\cite{meshlab}, to compute normal for the reconstructed grid edge points (GEP). The number of neighboring point clouds required for merging when calculating normal is set to the default value of 10. Subsequently, we use screened Poisson surface reconstruction (SPSR)~\cite{SPSR} for meshing the point clouds. Within SPSR, we adhere to several default parameters: the maximum depth of the octree used for reconstruction is set at 8, the minimum number of samples within each octree node is 1.5, and the interpolation weight between point clouds is 4. It is worth emphasizing that all models presented in our paper strictly follow these default VCGlib and SPSR settings, with no additional modifications post-reconstruction. This consistent use of default parameters ensures a standardized and reproducible approach across all our models.

%------------------%
\subsection{Configurations of Other Methods}

\noindent \textbf{SIREN}~\cite{SIREN}. We configure SIREN following the original paper. Notably, in our experimental setup, we deliberately abstain from utilizing the normal information of SIREN for supervision. This exclusion aims to maintain uniform supervision conditions across all methods.
Furthermore, our loss setup incorporates the Eikonal regularizer, denoted as $\mathcal \nabla|f| = 1$. This inclusion is specifically designed to align with our $\mathcal{L}_{der}$ term, thereby maintaining fairness and consistency in the comparative analysis of different methods.

\vspace{1mm}
\noindent \textbf{NGLOG}~\cite{nglod}. The official implementation of NGLOD in accordance with the original paper is adopted. To attain a more precise reconstruction model, we opt for a dense voxel version of NGLOD. This variation permits NGLOD to encompass global SDF details, resulting in a more complete surface reconstruction, thereby facilitating a fairer comparison among different reconstruction methods.

\vspace{1mm}
\noindent \textbf{NDF}~\cite{NDF}. While the original NDF is designed to learn and reconstruct a class of objects, including unseen ones, modifications to NDF are needed to align with single object reconstruction. To this end, we replicate 20 instances of a single training model as input. Since the input point clouds of each model consist of 50k points near the surface, our configuration yields approximately 1 million sample training points, ensuring a similar scale to the sample inputs used for UODFs. Moreover, to make a fair reconstruction comparison with UODFs, the point cloud estimated from NDF prediction also undergoes SPSR with masking. This revision is non-trivial as it bypasses the use of BPA~\cite{BPA}, whose parameter configuration is tailored to different models. 

\vspace{1mm}
\noindent \textbf{GIFS}~\cite{GIFS}. Similar to NDF's adjustment, GIFS is adapted for single object reconstruction. The training process involves replicating a single model 20 times to form a class of identical objects. This approach generates 1 million learning points per object. For mesh reconstruction, we employ GIFS' default mesh post-processing extraction algorithm, ensuring consistency with the original paper.

\vspace{1mm}
\noindent \textbf{HSDF}~\cite{HSDF}. HSDF also undergoes modifications to focus on learning and reconstructing a single model. Similar to NDF and GIFS, the sampling entails repeating a single model 20 times, resulting in a mesh model extraction using the masked marching cube (MC) algorithm proposed by HSDF.

%------------------%
\subsection{Parameter Amount}

Table~\ref{tab:paraAmount} presents a comprehensive comparison of the total number of parameters utilized in various methods. In our method, we employ a configuration of three $10$$*$$256$ and $3$$*$$256$ MLPs to separately regress UODF in each orthogonal direction. Despite this seemingly large network setup, our parameter amount (1.84M) is notably smaller than those in three UDF regression methods and the state-of-the-art SDF regression method NGLOD5.

\begin{table}[h]
    \begin{center}
        \caption{Comparison of parameter amount.}
        \label{tab:paraAmount}
        \vspace{-2mm}
        \resizebox{.48\textwidth}{!}{
        \begin{tabular}{|l||c|c|c||c|c|c||c|}
            \hline
            \multirow{2}{*}{Method} & \multicolumn{3}{c||}{SDF} & \multicolumn{3}{c||}{UDF} & \multicolumn{1}{c|}{UODFs} \\
            \cline{2-8} & \tabincell{c}{SIREN \\ \cite{SIREN}} & \tabincell{c}{NGLOD3 \\ \cite{nglod}} & \tabincell{c}{NGLOD5 \\ \cite{nglod}} & \tabincell{c}{NDF \\ \cite{NDF}} & \tabincell{c}{HSDF \\ \cite{HSDF}} & \tabincell{c}{GIFS \\ \cite{GIFS}} & \multicolumn{1}{c|}{\textbf{Ours}} \\
            \hline
            Paras. (M) & 0.199 & 0.199 & 10.1 & 4.62 & 6.60 & 3.68 & 1.84 \\
            \hline
        \end{tabular}
        }
    \end{center}
    \vspace{-8mm}
\end{table}

\section{Unstable Cases of SDF Sign}
\label{sec:supple_unstable}

SDF presents unique challenges in determining the sign at each sampling position, especially for complex shapes, unlike UDF and UODFs that only require specific distance values. This section discusses the instability issues associated with SDF computation using the following two categories of approaches.
%\vspace{-1mm}

\begin{itemize}
    \vspace{-1mm}
    \item \textit{First Category:} Implemented in the \textit{libigl} library~\cite{libigl}, used by NI~\cite{NI} and HSDF~\cite{HSDF}. \textit{libigl} employs the generalized winding number to determine the inside or outside positioning of points relative to a given mesh. \vspace{-1mm}
    \item \textit{Second Category:} Adopted in NGLOD~\cite{nglod} and Instant-NGP~\cite{instant}. NGLOD initially computes the unsigned distance, then determines the sign through ray stabbing in $13$ pre-defined directions. Instant-NGP follows a similar approach but uses BVH acceleration and ray stabbing with $32$ uniformly distributed directions.
    \vspace{-1mm}
\end{itemize}

\begin{figure}[t]
  \centering
  \includegraphics[width=\linewidth]{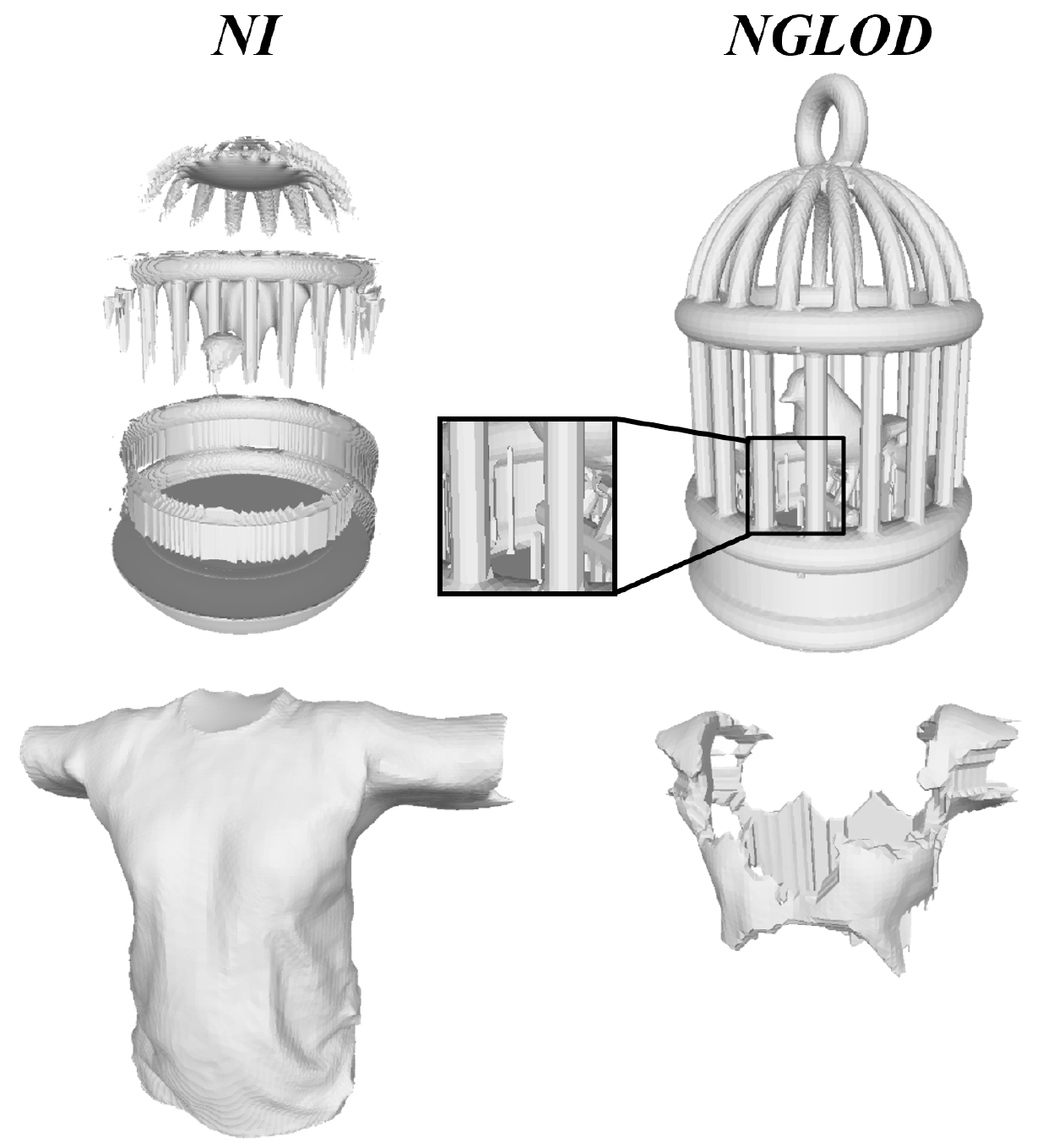}
  \vspace{-6mm}
  \caption{Reconstructions of the watertight birdcage and the non-watertight garment for two choices of SDF computation. Due to the unstable SDF sign, most of the shape surface may not be reconstructed.}
  \label{fig:unstableSDF}
  \vspace{-2mm}
\end{figure}

Figure~\ref{fig:unstableSDF} illustrates the marching cubes (MC) reconstruction results based on the ground truth SDF computed by NI and NGLOD, highlighting noticeable artifacts and detail errors. In contrast, UODFs do not exhibit these issues in ground truth computation, as evidenced by the comparative reconstruction metrics presented in Table~\ref{tab:birdcage}. The table reveals that MC reconstructions based on ground truth SDF are significantly less accurate and NGLOD tends to diminish high-frequency details in its regression to the ground truth SDF.
\vspace{-1mm}
\begin{table}[h]
\begin{center}
\caption{Comparison of reconstruction for the birdcage.}
\label{tab:birdcage}
\vspace{-3mm}
\resizebox{0.48\textwidth}{!}{
\begin{tabular}{|l||c|c|c|c|}
\hline
\diagbox{Metric}{Method} & NGLOD3 & NGLOD5 & SDF GT & \tabincell{c}{UODFs\\ \textbf{Ours}} \\
\hline
CD-GEP(*$10^5$)~$\bm{\downarrow}$ & 9.91 & 23.7 & 32.7 & \textbf{1.62} \\
\hline 
CD-Mesh(*$10^5$)~$\bm{\downarrow}$ & 14.0 & 27.7 & 39.2 & \textbf{5.90}\\
\hline
NC-Mesh~$\bm{\uparrow}$ & 95.9 & 94.5 & 93.1 & \textbf{97.7}  \\
\hline
\end{tabular}
}
\end{center}
\vspace{-6mm}
\end{table}

%------------------%
\section{Additional Details of UODFs Fitting and Reconstruction}
\label{sec:supple_fitting}

%------------------%
\subsection{2D UODF Results}

In this section, we present the additional details of UODFs fitting and reconstruction using two distinct instances: the closed `Rabbit' from the Thingi32 dataset and the `Dragon' from the Stanford Scanning dataset.
%Since UODF along one orthogonal direction is regressed by a constant MLP network, the fitting error of each UODF may differ obviously and should be analyzed individually.

\begin{figure*}[th]
  \centering
  \includegraphics[width=\linewidth]{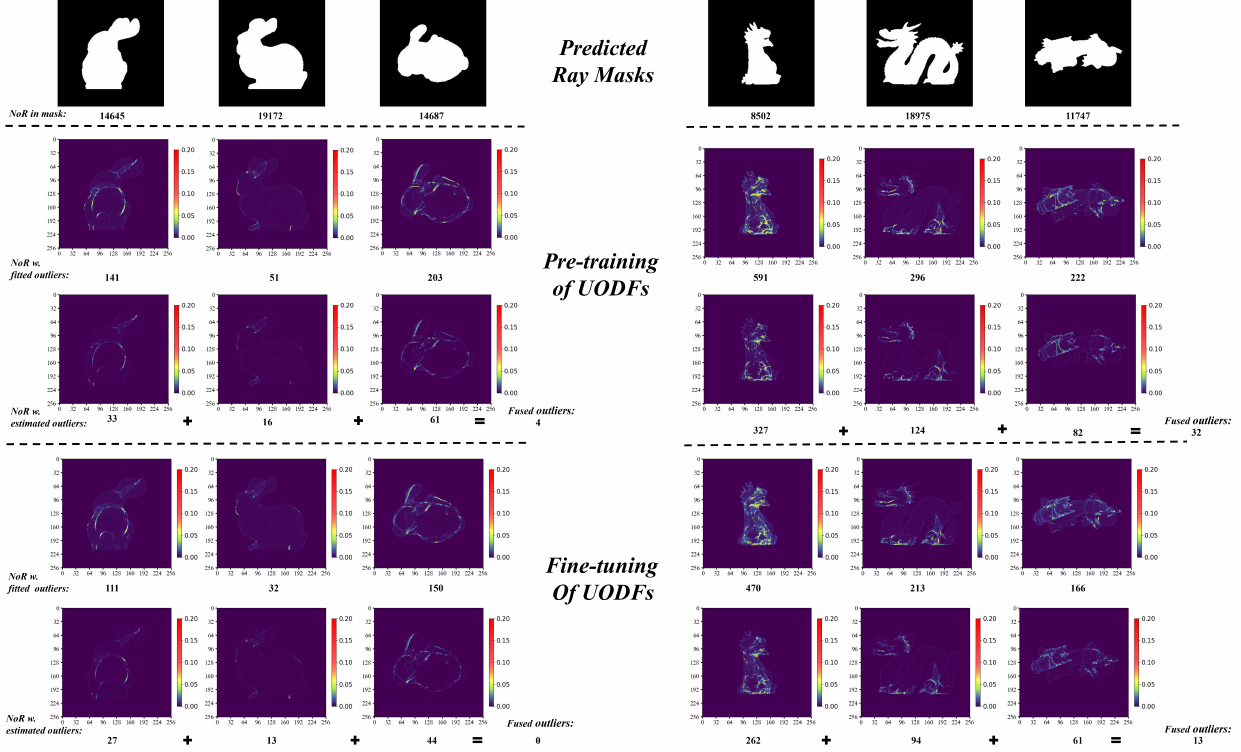}
  \vspace{-5mm}
  \caption{Detailed results of UODF fitting and reconstruction of grid edge points for `Rabbit' and `Dragon'.}
  \label{fig:fittingDetails_bunny}
  \vspace{2mm}
\end{figure*}

\vspace{2mm}
\noindent \textbf{Rabbit Instance}:
\vspace{-1mm}

\begin{itemize}
    \vspace{-1mm}
    \item The fitting and reconstruction results at $257^3$ grid corners are displayed in Fig.~\ref{fig:fittingDetails_bunny}. The first row depicts the predicted masks on three orthogonal planes, accompanied by the number of rays (NoR) in each mask.
    \vspace{-1mm}
    \item The subsequent row details the maximum errors in fitted UODF values along each orthogonal direction, along with the number of outliers (defined as points deviating more than 5 grids from the shape surface).
    \vspace{-1mm}
    \item A significant reduction of outliers is observed from the third row, due to averaging operations in estimating GEP, as discussed in Sec.~\textcolor{red}{3.5}. Moreover, the fusion operation results in only 4 final outliers, illustrating that most outliers in one orthogonal direction are isolated.
    \vspace{-1mm}
    \item Fine-tuning efforts can be seen in the last two rows, involving over-sampling rays with high prediction errors and additional training epochs. This leads to a noticeable decrease in outliers of the fitted UODF values, the estimated GEP, and the fused GEP.  
\end{itemize}

\begin{figure*}[t]
  \centering
    \includegraphics[width=\linewidth]{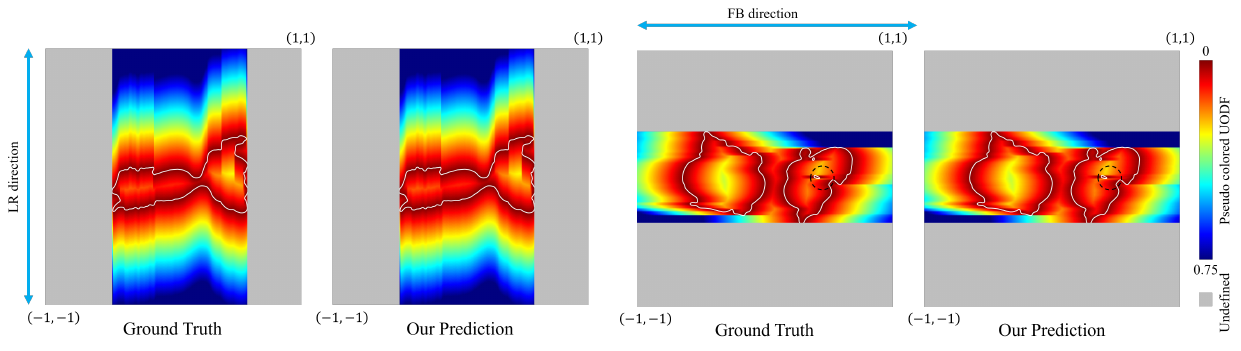}
  \vspace{-6mm}
  \caption{One slice of the $U\!O\!D\!F_{L\!R}$ and $U\!O\!D\!F_{F\!B}$ of the `Dragon' model. A subtle difference between ground truth and our prediction can be seen in the areas drawn by black dotted circles in the $U\!O\!D\!F_{F\!B}$ (with zoomed-in views).}
\label{fig:2Dfitting_Dragon_otherdirections}
\end{figure*}

The right part of Fig.~\ref{fig:fittingDetails_bunny} illustrates the results for the `Dragon' instance, being more complex. Consequently, the second to fifth rows show higher numbers of rays with outliers. Similarly, the number of final outliers in fused points reduced from 32 to 13 after fine-tuning, affirming the effectiveness of our approach. As the complement of Fig.~\textcolor{red}{3} in the main manuscript, Fig.~\ref{fig:2Dfitting_Dragon_otherdirections} shows a slice of UODF in the other two orthogonal directions for the `Dragon' model. There is a subtle difference between ground truth and our prediction, highlighted by black dotted circles.

\vspace{1mm}
These instances exemplify the effectiveness of our UODFs fitting and reconstruction process. \textbf{We emphasize that the experimental results in our manuscript and video are free of fine-tuning, although this can further improve the visual effect and accuracy metrics.}

%\vspace{2mm}

%------------------%
\section{Ablation of Point Merging Threshold}

\begin{table*}[t]
\begin{center}
\caption{Ablation study of the point merging threshold $\tau$. The metrics in each cell are CD-GEP(*$10^5$)~$\bm{\downarrow}$, CD-Mesh(*$10^5$)~$\bm{\downarrow}$, and NC-Mesh~$\bm{\uparrow}$, respectively.}
\label{tab:ablation_threshold}
%\vspace{-2mm}
\resizebox{0.9\textwidth}{!}{
\begin{tabular}{|l||c|c|c|c|c|}
\hline
\diagbox{$\tau$}{Shape} & `Nandi the Bull' & `Hilbert Cube' & Lamp & Birdcage & `Armadillo' \\
\hline
1/512 & \small{0.228 / 2.77 / 98.5} & \small{0.00532/12.0/92.1} & \small{0.0399/2.43/98.8} & \small{1.62/5.90/97.7}  & \small{0.684/2.37/97.8}  \\
\hline 
1/256 & \small{0.217 / 2.77 / 98.5} & \small{0.00523/12.0/92.0} & \small{0.0398/2.43/98.7} & \small{1.66/5.93/97.6}  & \small{0.726/2.39/97.8}  \\
\hline
1/128 & \small{0.210 / 2.76 / 98.5} & \small{0.00523/12.0/92.0} & \small{0.0405/2.44/98.8} & \small{1.69/6.02/97.7}  & \small{0.707/2.39/97.8}  \\
\hline
\end{tabular}
}
\end{center}
\vspace{-4mm}
\end{table*}

This section presents an ablation study on the point merging threshold $\tau$, initially introduced in Sec.~\textcolor{red}{3.5}. While the threshold $\tau$ is set to $1/512$ (a quarter of the grid size at a $256^3$ resolution) in our primary experiments, we explore its impact on reconstruction accuracy by adjusting its value.
$\tau$ is additionally set to $1/256$ (half of the grid size) and $1/128$ (equal to the grid size). Along with the `Nandi the Bull’ and `Hilbert Cube’ from the main manuscript, we include the `Lamp' from ShapeNet, the `Birdcage' from Thingi10K, and the `Armadillo' from the Stanford dataset.

Table~\ref{tab:ablation_threshold} shows the three metrics across five shapes. It indicates that the reconstruction accuracy remains largely unaffected by variations in the point merging threshold. Even at a threshold of $1/128$, the accuracy metrics show minimal sensitivity to this parameter. This finding suggests a robustness in our method's performance relative to the point merging threshold.

\section{Additional Reconstruction Results}

\begin{table*}[t]
\renewcommand\arraystretch{1.1}
    \begin{center}
        \caption{Detailed reconstruction metrics for each of ten garments from the MGN dataset. The two CD metrics~$\bm{\downarrow}$ are multiplied by a factor of $10^5$.}
        \label{tab:MGNDataset}
        %\vspace{-4mm}
        \resizebox{1.0\textwidth}{!}{
        \begin{tabular}{|l||c|c|c|c|c|c|c|c|c||c|c|c|}
            \hline
            \multirow{2}{*}{Method} & \multicolumn{9}{c||}{UDF regression} & \multicolumn{3}{c|}{UODFs regression} \\
            \cline{2-13} & \multicolumn{3}{c|}{NDF~\cite{NDF}} & \multicolumn{3}{c|}{HSDF~\cite{HSDF}} & \multicolumn{3}{c||}{GIFS~\cite{GIFS}} & \multicolumn{3}{c|}{\textbf{Ours}} \\
            \hline
            {\diagbox{Shape}{Metric}} & \tabincell{c}{CD-\\GEP} & \tabincell{c}{CD-\\Mesh} & \tabincell{c}{NC-\\Mesh} & \tabincell{c}{CD-\\GEP} & \tabincell{c}{CD-\\Mesh} & \tabincell{c}{NC-\\Mesh} & \tabincell{c}{CD-\\GEP} & \tabincell{c}{CD-\\Mesh} & \tabincell{c}{NC-\\Mesh} & \tabincell{c}{CD-\\GEP} & \tabincell{c}{CD-\\Mesh} & \tabincell{c}{NC-\\Mesh} \\
            \hline
            TshirtNoCoat-1 & 146 & 145 & 94.4 & 8.52 & 10.1 & 90.0 & 2.13 & 3.32 & 97.7 & \textbf{0.212} & \textbf{1.87} & \textbf{99.8} \\
            \hline
            ShortPants-1 & 154 & 155 & 93.6 & 13.5 & 17.7 & 93.4 & 2.04 & 5.05 & 97.4 & \textbf{0.162} & \textbf{3.60} & \textbf{99.6} \\
            \hline
            Pants-1 & 81.7 & 82.0 & 94.0 & 14.4 & 15.8 & 90.1 & 2.0 & 3.49 & 95.4 & \textbf{0.284} & \textbf{2.04} & \textbf{99.6} \\
            \hline
            Pants-2 & 148 & 145 & 91.5 & 17.9 & 19.4 & 88.3 & 3.78 & 5.07 & 90.9 & \textbf{0.182} & \textbf{2.05} & \textbf{99.7} \\
            \hline
            ShirtNoCoat-1 & 34.9 & 34.8 & 93.7 & 12.0 & 11.8 & 81.7 & 16.5 & 16.1 & 84.6 & \textbf{0.311} & \textbf{0.929} & \textbf{99.7} \\
            \hline
            TshirtNoCoat-2 & 158 & 156 & 91.5 & 7.98 & 9.40 & 91.0 & 2.12 & 3.35 & 95.6 & \textbf{0.156} & \textbf{1.86} & \textbf{99.8} \\
            \hline
            ShortPants-2 & 245 & 248 & 93.4 & 13.2 & 17.2 & 91.6 & 2.46 & 5.21 & 97.3 & \textbf{0.143} & \textbf{3.43} & \textbf{99.7} \\
            \hline
            ShirtNoCoat-2 & 107 & 104 & 77.2 & 13.6 & 13.1 & 85.2 & 8.24 & 8.0 & 86.7 & \textbf{0.183} & \textbf{0.777} & \textbf{99.7} \\
            \hline
            LongCoat-1 & 135 & 141 & 88.1 & 10.4 & 11.4 & 88.4 & 6.62 & 7.24 & 86.0 & \textbf{0.368} & \textbf{1.69} & \textbf{98.9} \\
            \hline
            LongCoat-2 & 66.9 & 65.0 & 90.7 & 18.0 & 18.3 & 80.5 & 3.61 & 3.88 & 88.5 & \textbf{0.264} & \textbf{1.07} & \textbf{99.7} \\
            \hline
        \end{tabular}
        }
    \end{center}
\end{table*}

This section presents additional reconstruction results not included in the main manuscript.

\begin{figure*}[t]
  \centering
    \includegraphics[width=\linewidth]{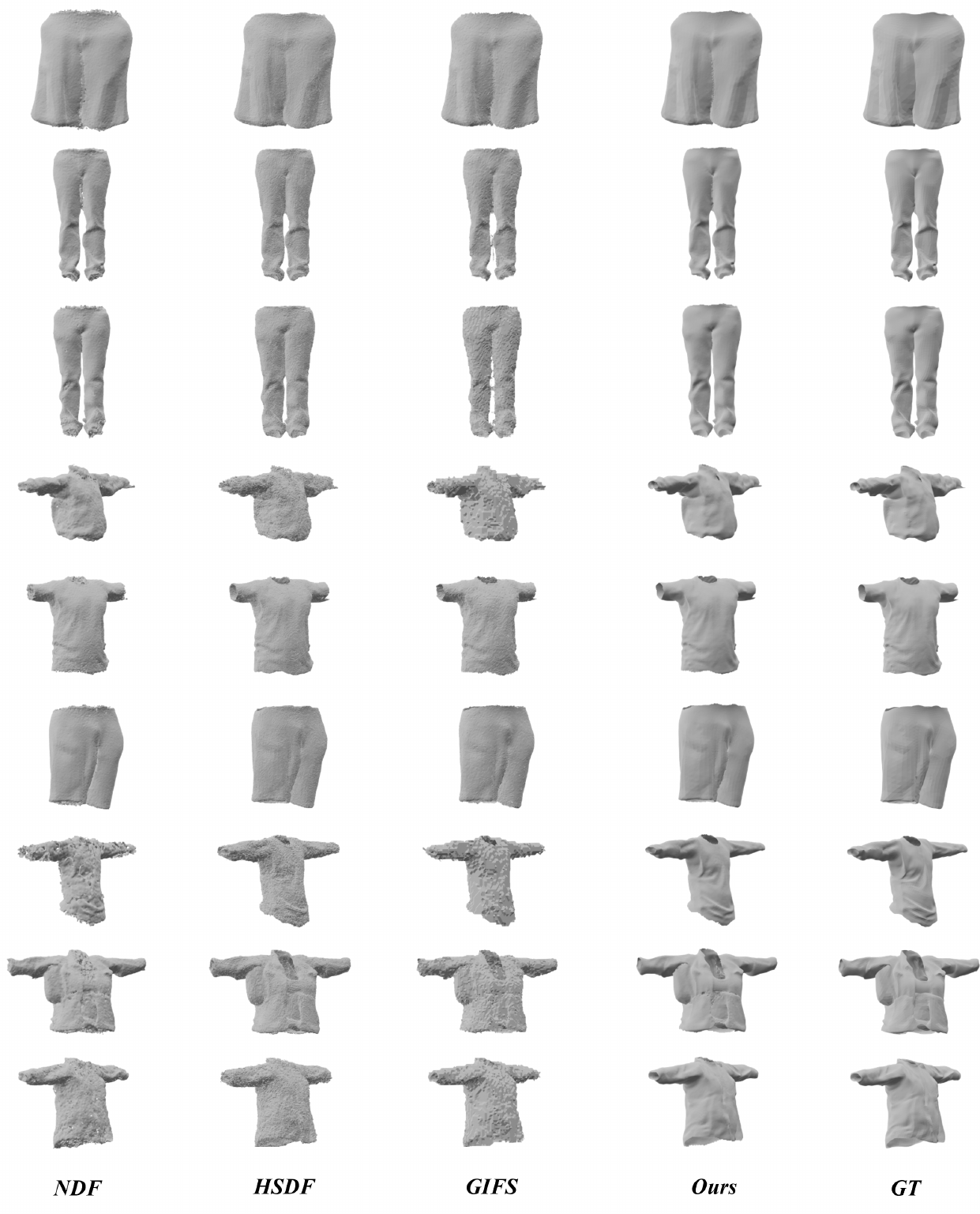}
  \vspace{-6mm}
  \caption{Reconstruction results for other nine shapes in the MGN10 dataset.}
  \label{fig:additionalMGN}
\end{figure*}

First, we elaborate on the three reconstruction metrics for the ten non-watertight garments from the MGN dataset in Table~\ref{tab:MGNDataset}. These results show that our UODFs based NIR method significantly outperforms three other UDF regression methods. We have already showcased the `TshirtNoCoat-1' garment in Fig.~\textcolor{red}{9} in the main manuscript and the results for the remaining nine garments are illustrated in Fig.~\ref{fig:additionalMGN}.

\begin{figure*}[t]
  \centering
  \includegraphics[width=\linewidth]{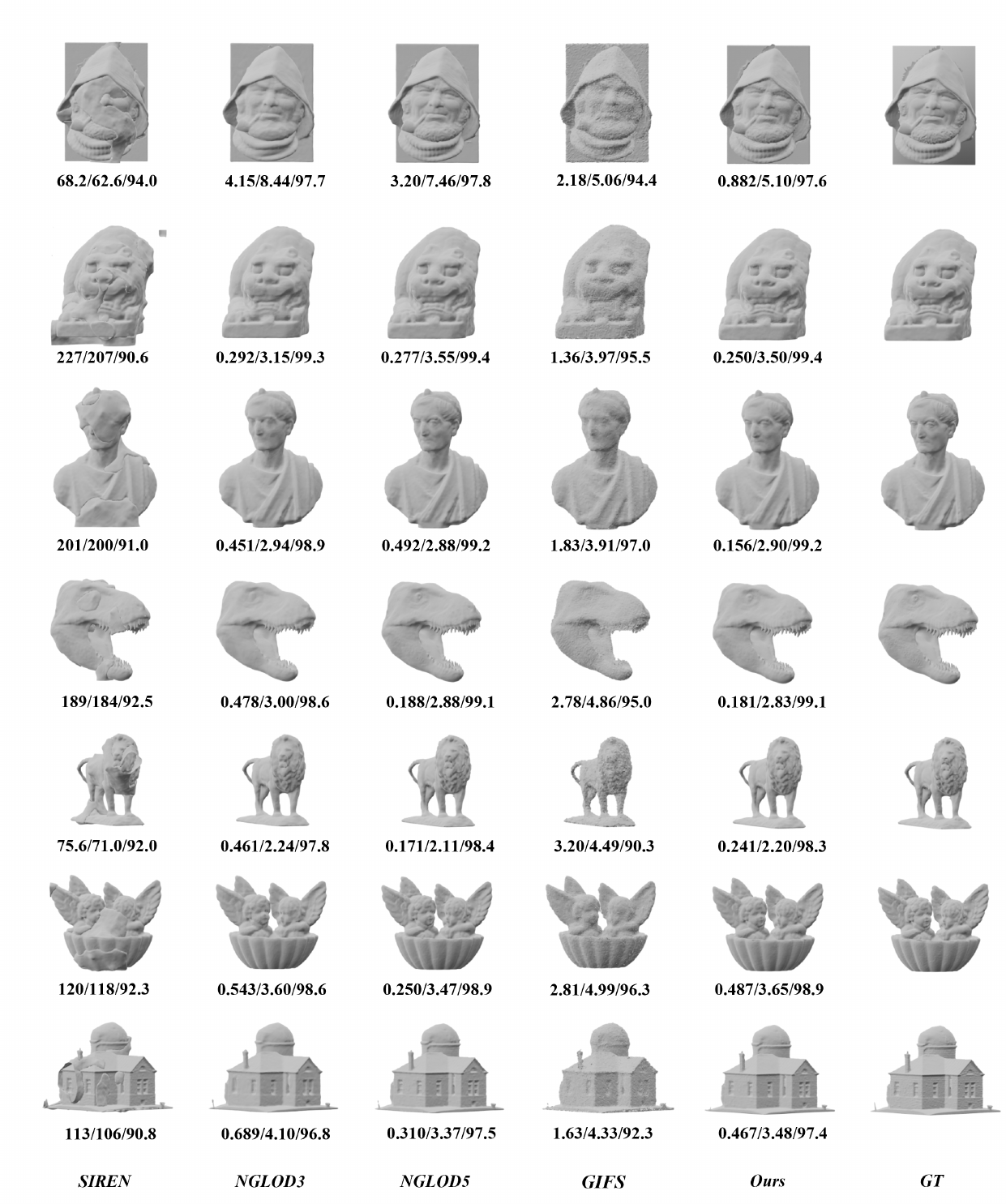}
  \vspace{-6mm}
  \caption{Reconstruction results for more shapes in the Thingi32 dataset. The metrics from left to right below each shape are CD-GEP(*$10^5$)~$\bm{\downarrow}$, CD-Mesh(*$10^5$)~$\bm{\downarrow}$, and NC-Mesh~$\bm{\uparrow}$, respectively.}
  \label{fig:additionalThingi32}
\end{figure*}

Second, for the watertight shapes, additional results on the Thingi32 dataset are displayed in Fig.~\ref{fig:additionalThingi32}, demonstrating consistency with the findings shown in Fig.~\textcolor{red}{6} and Table~\textcolor{red}{1} in the main manuscript.

\begin{figure*}[t]
  \centering
    \includegraphics[width=\linewidth]{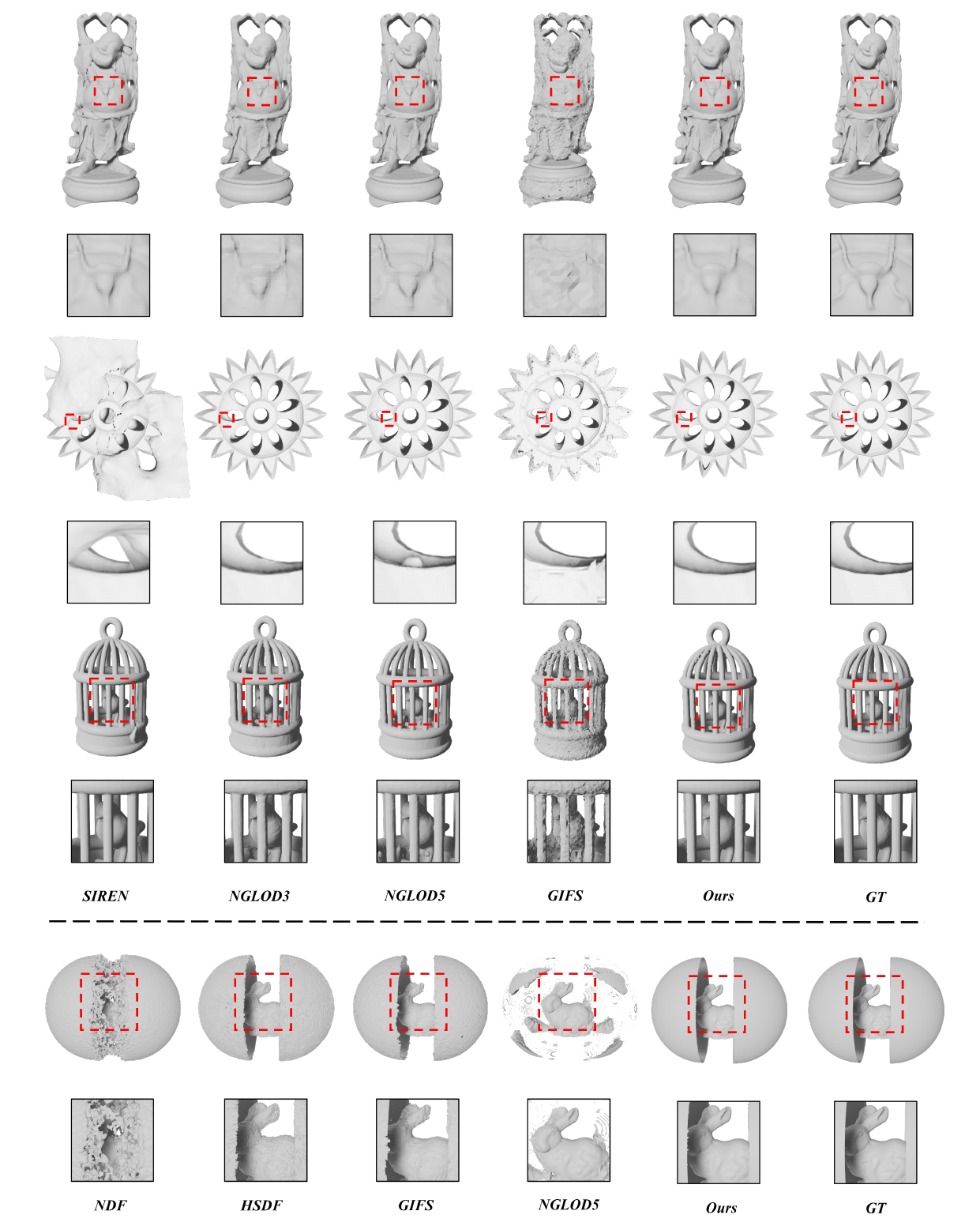}
  \vspace{-6mm}
  \caption{Reconstruction results for additional four complex shapes.}
  \label{fig:additionalComplex}
\end{figure*}

Third, we present the outcomes for the additional four complex shapes in Fig.~\ref{fig:additionalComplex}. The zoomed-in views further underscore the high-quality shape reconstruction achievable with our proposed UODFs.

\begin{figure*}[t]
  \centering
    \includegraphics[width=\linewidth]{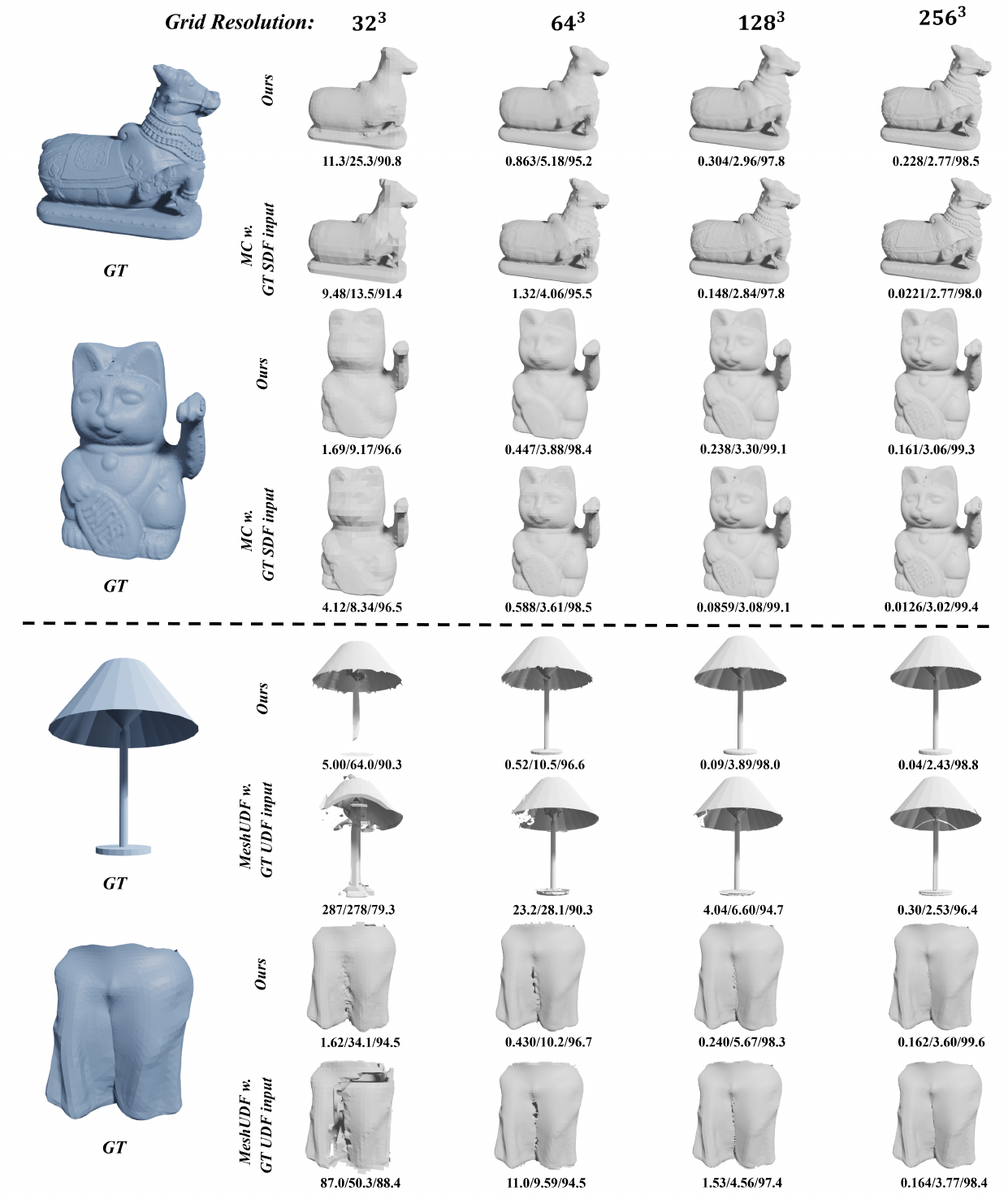}
  \vspace{-6mm}
  \caption{Additional reconstruction results at various resolutions of grids. The metrics from left to right below each shape are CD-GEP(*$10^5$)~$\bm{\downarrow}$, CD-Mesh(*$10^5$)~$\bm{\downarrow}$, and NC-Mesh~$\bm{\uparrow}$, respectively.}
  \label{fig:additionResolution}
\end{figure*}

Finally, more reconstruction experiments at various grid resolutions from $32^3$ to $256^3$ are conducted. Fig.~\ref{fig:additionResolution} shows the results of four distinct shapes: two watertight and two non-watertight. To highlight the reconstruction accuracy of our UODF-based NIR method, we compare it with the marching cube (MC) algorithm~\cite{Marching_cubes} and MeshUDF~\cite{meshUDF}, that utilize ground truth SDF and UDF as input, respectively. Therefore, the MC and MeshUDF results do not suffer from fitting errors of neural network, but are only affected by interpolation errors. For these two watertight models, our method achieves performance comparable to the MC method across all grid resolutions. In the case of non-watertight models, our method significantly outperforms MeshUDF using the ground truth UDF as input, at all grid resolutions. MeshUDF has some spurious meshes at the edge of shapes, especially under lower resolutions. In addition, our meshes generated by SPSR~\cite{SPSR} are smoother than those extracted by MC and MeshUDF.

% \newpage
% \vspace{10mm}
%%%%%%%%% REFERENCES
% {\small
% \bibliographystyle{ieee_fullname}
% \bibliography{my}
% }

\newpage

\newpage

\newpage

\newpage

\end{document}